# An Image Processing based Object Counting Approach for Machine Vision Application

Mehmet BAYGIN[1], Mehmet KARAKOSE[2]
Computer Engineering Department
[1]Ardahan University, 75000
Ardahan, Turkey
mehmetbaygin@ardahan.edu.tr, mkarakose@firat.edu.tr

Alisan SARIMADEN[3], Erhan AKIN[2]
Computer Engineering Department
[2]Firat University, 23119
Elazig, Turkey
medel@medelelektronik.com, eakin@firat.edu.tr

*Abstract*— Machine vision applications are low cost and high precision measurement systems which are frequently used in production lines. With these systems that provide contactless control and measurement, production facilities are able to reach high production numbers without errors. Machine vision operations such as product counting, error control, dimension measurement can be performed through a camera. In this paper, a machine vision application is proposed, which can perform object-independent product counting. The proposed approach is based on Otsu thresholding and Hough transformation and performs automatic counting independently of product type and color. Basically one camera is used in the system. Through this camera, an image of the products passing through a conveyor is taken and various image processing algorithms are applied to these images. In this approach using images obtained from a real experimental setup, a real-time machine vision application was installed. As a result of the experimental studies performed, it has been determined that the proposed approach gives fast, accurate and reliable results.

## I. INTRODUCTION

Production lines have become very easily controllable and auditable with machine vision applications [1, 2]. With these systems measurement operations can be carried out without the need for an expert control. These measurements are made in a completely non-contact manner, and defective products are sorted out as they pass through the conveyor [3]. Very fast machine vision systems allow for many measurements in a short period of time. This increases the instantaneous production capacity of production facilities. It is critical that machine vision based applications are able to work with near perfect precision [4, 5]. The products to be delivered to the end user must be complete, exact and identical, and are the most important criteria in these applications [6].

A conveyor is basically a system in which products are passed over and faulty products are sorted out on this band. An image of the products through a camera integrated in this system is taken and the faulty products are determined [7]. The general structure of these systems is shown in Fig. 1.

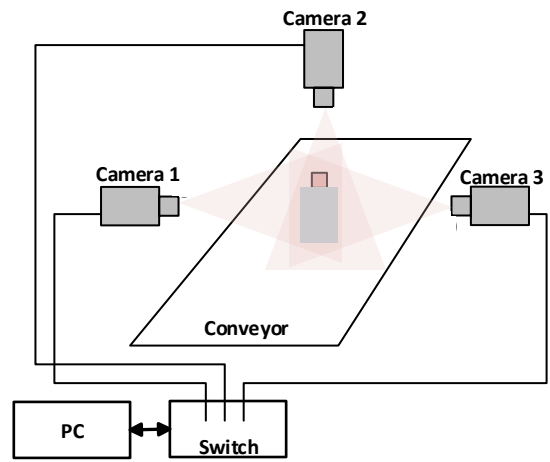

Fig. 1. General structure of machine vision systems [7]

Object counting is a frequently encountered problem in many machine vision applications [8]. Especially in production systems operating at high speed, it is very difficult to perform measurement without using computer vision. Object counting is a frequently encountered problem in many machine vision applications. Especially in production systems operating at high speed, it is very difficult to perform measurement without using computer vision. As a result, the use of machine vision systems becomes essential. In one of these studies using high speed cameras, products passing through the conveyor were recorded via a camera capable of recording 60 fps. In this study, basement background extraction was performed and moving objects were detected. Then, it is checked whether the objects in a specified area pass or not, and the counting process is performed. A block diagram summarizing this proposed approach is as shown in Fig. 2 [9].

Image processing based computer vision applications use many different areas. In one of the works carried out for this purpose, image processing is used for diagnosing the fault. In the proposed method, pantograph catenary images taken from the railway are used. Images were thresholded using the otsu method. The obtained results were combined with the time series and particle swarm optimization to detect faults. The flow diagram of the proposed method is given in Fig. 3. [10].

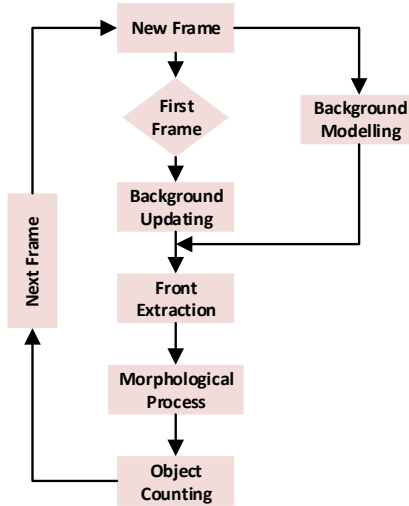

Fig. 2. A proposed approach in the literature [9]

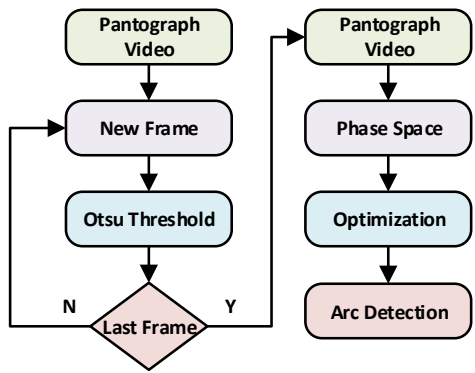

Fig. 3. A proposed approach in the literature [10]

In a study related to the subject, an application was developed to detect biscuit cracks for quality control. In the study, the characteristics of the cracks were determined using Hough transformation, Canny filter and Laplace edge extraction [11]. In another study, object counting was performed. In this application, which is performed for counting the objects which are overlaid, the boundaries of the objects are determined and the counting process is performed with classical geometric operations [12]. Khlule and dig. [13] performs the counting of the objects contained within the tablets. The proposed approach includes the steps of input image, preprocessing, segmentation and counting. The automatic counting method proposed in this study is tested on two applications and high performance is achieved.

In this study, an automatic object counting method based on grassy thresholding has been developed. Images of products passing through a moving conveyor were taken through a camera and pre-processed. Later, Otsu and Hough methods were applied to these images, and the detection and counting of the products was performed. Eggs and soda bottles were used in the tests and high precision counting process was performed**.**

## II. Proposed Method

The object counting process that is frequently encountered in machine vision applications are used in many areas. Detection of objects on conveyors running at high speed and counting with high precision is a very important issue. The counting of the objects in the egg and soda bottle packets performed with the proposed approach in this study. The proposed approach works independently of the object type in real time. Sample images of the objects used in the study are shown in Fig. 4.

The study mainly involves the monitoring, detection and counting of the products passing through the conveyor by means of a camera. For this purpose, Otsu thresholding and Hough transformations were used. The images used in the study were obtained from a system that operates in real time. A camera that monitors the conveyor from above is integrated into the system and the images of the products are taken instantaneously. The steps of the machine vision based automated product counting approach proposed in this study are presented in Fig. 5.

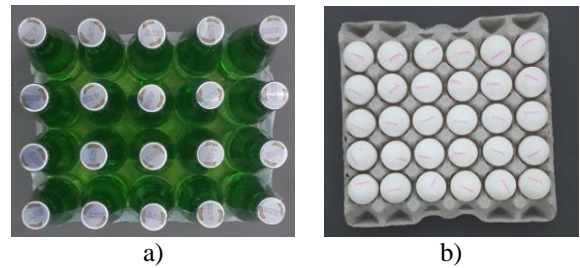

a) b)

Fig. 4. Sample images a) Soda bottle pack b) Egg packet

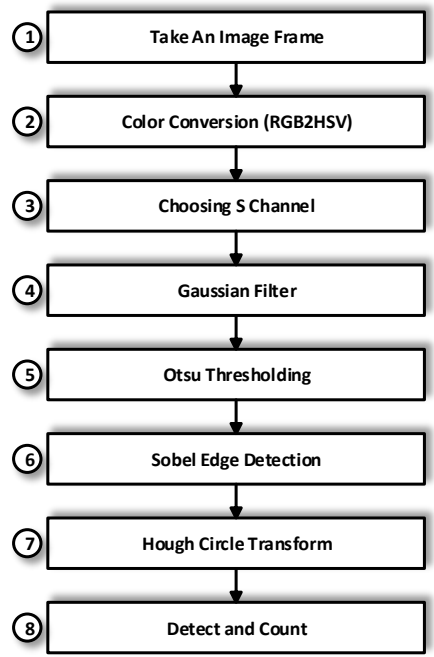

Fig. 5. A block diagram of proposed method

The proposed approach first takes an image of the object through the camera. The image in RGB color format is converted into HSV color space. The aim of this transformation is to be able to obtain the details of the image more clearly, independent of the light conditions [14]. Only the saturation (S) channel is used in the system after this conversion. The conversion from RGB color space to HSV color space and the calculation of S channel are given in Equation 1.

$$MAX = \max\{R, G, B\}$$
$$MIN = \min\{R, G, B\}$$
$$S = \begin{cases} 0, & if \ MAX = 0 \\ 1 - \frac{MIN}{MAX}, & else \end{cases} \quad (1)$$

After reducing the image to single channel, this image is subjected to a Gaussian filter. The general mathematical expression of this filter used to eliminate the noises on the image is presented in Equation 2.

$$G(r) = \frac{1}{(2\pi\sigma^2)^{N/2}} e^{-r^2/(2\sigma^2)} \quad (2)$$

The threshold values in the image can be determined adaptively by the Otsu threshold method. Techniques using this process in the literature generally work according to the two-level Otsu method [15]. Three-level Otsu method was developed and used in this study. With this approach, adaptive 3 different threshold levels are determined and the image applied by gauss filter is divided into 4 color ranges. In this process the aim is to reveal all the details in the image. In the other step of the proposed approach, edge detection of the thresholded image is performed. Sobel edge detection method is used in this process. The reason why this method is preferred is that it is very easy to apply. Basically, two convolution matrices, called Gx and Gy, are the process of navigating through the image data [16, 17]. These convolution matrices are given in Equation 3.

$$G_x = \begin{bmatrix} -1 & 0 & 1 \\ -2 & 0 & 2 \\ -1 & 0 & 1 \end{bmatrix}, \ G_y = \begin{bmatrix} 1 & 2 & 1 \\ 0 & 0 & 0 \\ -1 & -2 & -1 \end{bmatrix} \quad (3)$$

Hough transformation is used in the last step of the proposed method. This method basically works with the ratios of the geometric shapes of the edges [18, 19]. With this method applied to the image removed the edges, geometric objects in the image can be detected. A pseudo code fragment shows the flow of this method are given in Fig. 6.

The Hough transformation marks the geometric shapes identified as the working principle and finds the center of these shapes [20, 21]. In this sense, there is no need for a process such as determining the boundaries of objects again. In this case, the algorithm does not perform any extra processing and contributes to the algorithm in terms of time.

```
For each pixel (x,y)
    For each radius r=10 to r=60
        For each theta t=0 to 360
            a=x – r * cos(t * PI/180);
            a=x – r * cos(t * PI/180);
            A[a, b, r] + = 1;
        end
    end
end
```

Fig. 6. A pseudo code for Hough circle transform

III. EXPERIMENTAL RESULTS

The data obtained in this study was obtained from a real-time working system. For this purpose, a camera has been installed which continuously monitors the conveyor. This camera has a capacity of 60 frames per second. Imagery from the camera is instantly transferred to a computer via the Ethernet port. On the computer side of the system, an automatic object counting algorithm developed within the scope of the study is carried out. A block diagram summarizing the experimental setup presented is shown in Fig. 7. Detailed features such as resolution, frame rate of the camera used in the study are given in Table 1.

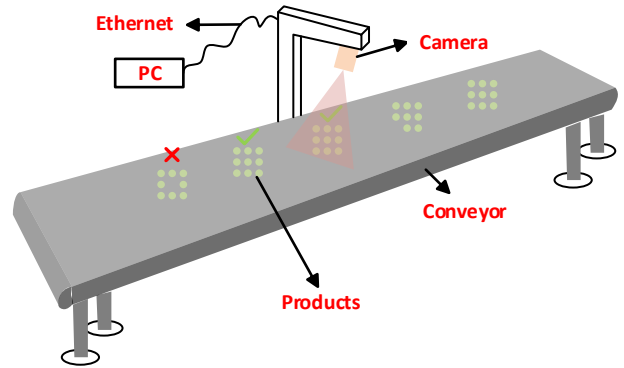

Fig. 7. A pseudo code for Hough circle transform

TABLE I. CAMERA FEATURES USED STUDY

| Camera Features | |
|---|---|
| Feature | Values |
| Resolution | 1280x720 px |
| Frame Rate | 59 fps |
| Shutter Type | Global |
| Sensor Type | CCD |
| Interface | GigE |
| Pixel Bit Depth | 12 bit |
| Mono/Color | Color |

In this study, the test procedures were performed for two different product groups. These products are egg and soda bottles respectively. The same procedure was applied to both test groups and successful results were obtained. The sample image processing results obtained by applying the proposed approach on egg packages are given in Fig. 8.

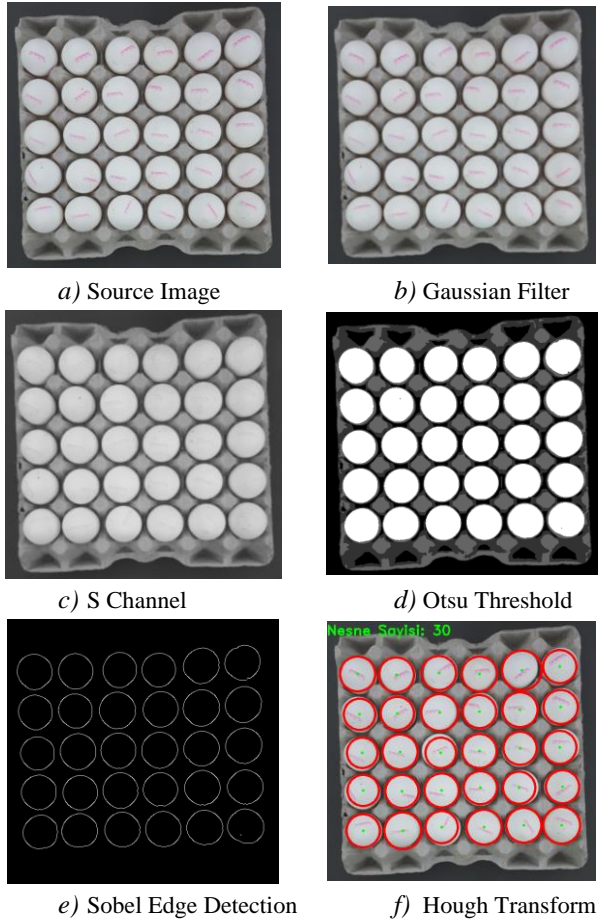

*a)* Source Image   *b)* Gaussian Filter
*c)* S Channel   *d)* Otsu Threshold
*e)* Sobel Edge Detection   *f)* Hough Transform
Fig. 8. The proposed automatic object counting method

As can be seen from Fig. 8, the method basically consists of 8 steps. In this frame, the egg packages were tested under different conditions. The sample results from these tests are shown in Fig. 9.

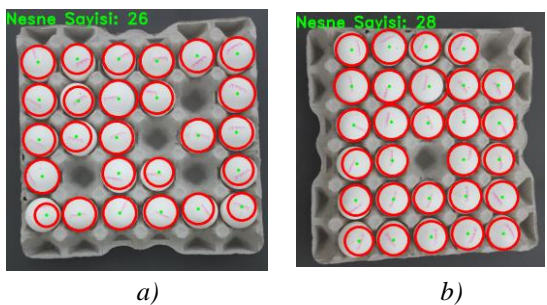

*a)*   *b)*
Fig. 9. The example test results for egg packets

As mentioned at the beginning of the section, the studies was carried out for two different product groups. One of these products is soda bottle packages. All of the steps of the image processing application are implemented in these images and good results are obtained. The results of the image processing algorithm performed for this product group are presented in Fig. 10.

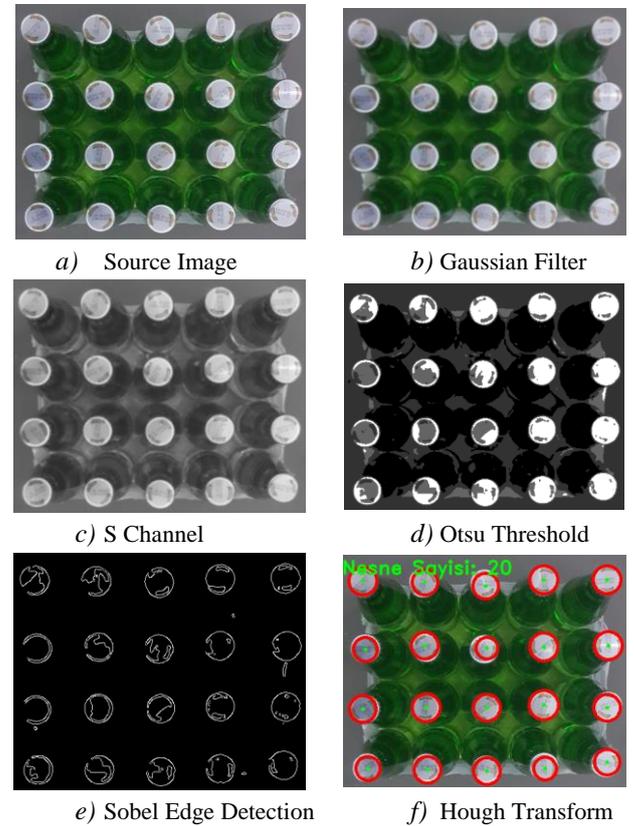

*a)* Source Image   *b)* Gaussian Filter
*c)* S Channel   *d)* Otsu Threshold
*e)* Sobel Edge Detection   *f)* Hough Transform
Fig. 10. Steps of the proposed approach

All the steps of the proposed approach are tested in this product group and very good results are obtained. As in the case of egg packages, test procedures were performed in different conditions within these products and the results of these tests are given in Fig. 11.

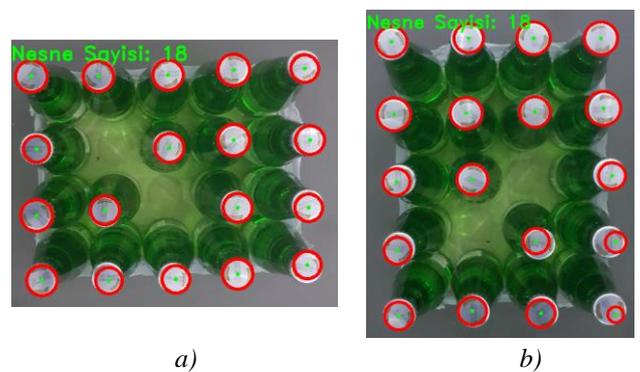

*a)*   *b)*
Fig. 11. The example test results for soda bottle packets

IV. CONCLUSIONS

Today, machine vision applications are actively used almost entirely in production systems. Especially contactless measurement is preferred because of its ability to provide high speed and precise measurement. Machine vision based systems have the some advantages such as low installation and maintenance costs as well as excellent production capacity of these systems.

In this paper, object counting, which is a frequently encountered problem in machine vision, has been performed. When the studies done in the literature are examined, it is seen that the applications carried out are directed to a specific product group. With the proposed approach, we have come from the top of this situation and have been able to count different product groups with a single algorithm. For this purpose, Otsu thresholding and Hough transformation methods are used. The images were taken in real time from a conveyor and tested with the machine vision application suggested in the study. When compared with the current studies in the literature, the proposed approach has shown that the method can measure accurately, quickly and with high accuracy.

ACKNOWLEDGMENT

This study has been supported by The Scientific and Technological Research Council of Turkey (SANTEZ Programme) under Research Project No: 0743.STZ.2014 (TUBITAK Grant No:112D021).